# Concept of a Sensor Test Environment for Dusty Agricultural Conditions

**Peter Buckel** [*,**] **Johannes Hermann** [**] **Jonas Wollmann** [**]
**Thomas Dietmüller** [**] **Timo Oksanen** [*]

[*] *Technical University of Munich, Chair of Agrimechatronics and Munich Institute of Robotics and Machine Intelligence (MIRMI), 85354 Freising, Bavaria, Germany. (e-mail: firstname.surname@tum.de)*
[**] *Baden-Wuerttemberg Cooperative State University (DHBW) Ravensburg, 88045 Friedrichshafen, Baden-Wuerttemberg, Germany. (e-mail: surname@dhbw-ravensburg.de)*

**Abstract:** Dust in agriculture presents a significant challenge for autonomous agricultural machinery. Dust can impair the performance of sensors and algorithms. This work, therefore, presents a concept for a proving ground consisting of an indoor and outdoor area. The indoor area comprises a laboratory test bench where dust circulates in a closed system and a test hall where life-size objects can be placed. The outdoor area features dedicated test setups that enable reproducible data to be recorded with and without dust during real-world agriculture work. The proving ground and the setups are visualized in 3D.

*Keywords:* Autonomous Agricultural Machinery, Proving Ground, Sensor Validation, Dust in Agriculture, Algorithm Development

## 1. INTRODUCTION

Sensor systems in agriculture are exposed to harsh environmental conditions such as haze, fog, rain, snow, and dust. According to ISO 18497-2 (2024), such environmental conditions can cause misclassification of objects. Thus, researchers proposed test benches to investigate environmental conditions in agriculture. Krause et al. (2022) proposed a test environment based on a rail-based carrier system with sensors mounted on top. The carrier system moves from arable to grassland, allowing reproducible recording over a long period at the same time of day and season. Meltebrink et al. (2021) proposed a setup to test and compare object detection systems of autonomous agricultural machinery. The setup was developed to evaluate the real environment detection areas of different sensors.

Buckel et al. (2024, 2023) demonstrated that it is possible to remove dust from images to improve high-level vision tasks. Buckel et al. (2023) recorded images with and without dust with a sensor mounted to a half-sided gate, and a tractor cultivating the field drives through it. Therefore, they validated the performance of state-of-the-art algorithms for dust removal. Additionally, Buckel et al. (2023) showed that dust removal can improve person detection. In another work, Buckel et al. (2024) proved that dust removal improves tillage quality monitoring.

In autonomous driving in road traffic, great efforts are being made to reproduce environmental conditions. Sensors and algorithms are commonly tested and validated in fog chambers (Bijelic et al., 2019; Rothmeier and Huber, 2021; Bijelic et al., 2018). A fog chamber is a closed hall where testing objects can be placed at various distances, and the fog density can be controlled (Colomb et al., 2004). The fog chamber enables detailed validation, e.g., how fog density impacts object detection performance. Although this type of test environment is widely used in the automotive sector, similar tests for agricultural machinery and sensor technology are not reported.

This work has identified the need for test setups for reproducible dusty environments to develop sensor systems and algorithms. Thus, a concept of a proving ground with various test setups is proposed. Overall, the main contributions are:

- A concept for a test site specially designed for replicating dusty real-world scenarios for testing purposes.
- Novel concepts of test setups enable year-round testing and development of sensors and algorithms in dusty agricultural conditions. The contributions of the respective concepts are:
  1. An indoor laboratory test bench with dust circulation in a closed-loop system for reproducible validation with a camera or stereo camera.
  2. An indoor test hall for year-round indoor testing with standardized real-world-sized objects.
  3. An outdoor gate-based test bench based on an aluminum travers with a rail system through which the tractor can drive while cultivating the soil is used to generate shift-free sensor data with and without dust during tillage.
  4. An outdoor cart-based setup attached to the implement to investigate object detectability in the center of the dust cloud.

## 2. REQUIREMENTS

### 2.1 Dust in Agriculture

Dust in agriculture is a particular challenge for autonomous agricultural machinery. Therefore, the first step is identifying where dust is produced in agriculture and where it can interfere with autonomous machines or robots. The following gives an overview of agricultural activities where the dust of some sort might affect machine operation and sensor systems:

- Soil-borne dust raised during
  - Tillage
  - Seeding
  - Hoeing
  - And more
- Straw dust raised during
  - Livestock farming, e.g., while spreading straw
  - During harvesting with a combine harvester
  - Straw or hay bale pressing
  - And more
- Fertilizer dust raised during
  - Spreading synthetic fertilizer
  - Spreading organic fertilizer
  - And more
- And more

The list highlights the wide range of areas in which dust is raised. However, straw dust may not influence environmental perception as much as soil-borne dust, for example. Nevertheless, straw dust can significantly impact specific monitoring applications; for example, the distribution of straw during harvesting. The authors think that the environmental perception might be most affected by dust raised during tillage and seeding due to the quantity and density of raised dust. Consequently, the proposed test environment and the different setups are tailored towards soil-borne dust raised during farming activities, rather than by wind.

### 2.2 General and Soil-borne Dust specific Requirements

Test setups must meet various requirements. On the one hand, there are general requirements that must be fulfilled, and on the other hand, there are dust-specific requirements.

*General Requirements:*

- Test setups shall enable reproducible data recording.
- All parameters and test conditions shall be documentable and recorded
- Test setups shall allow sensor data to be recorded with and without dust, allowing a thorough validation with dedicated metrics.
- A testing proofing ground should allow year-round testing of sensors and algorithms.
- The tests should be conducted at various levels, ranging from laboratory experiments and indoor tests involving full-size objects, to outdoor tests.
- The test setups should be designed to enable the investigation of object detection and tracking tasks in dusty conditions, as well as the general influence of dust on sensors such as LiDAR (Light Detection and Ranging), RADAR (Radio Detection and Ranging), single cameras and stereo cameras.

*Dust and Agriculture-specific Requirements:*

- Sensors should be appropriately protected against environmental influences, in particular dust.
- Real-world dust should be used in test setups to minimize the domain gap.
- The distribution and density of the dust, as well as the lighting conditions, should be realistic.
- To ensure the transferability of the results, dust must be stirred up during soil cultivation for outdoor test setups.

The requirements identified are based on the authors' experience.

## 3. TEST ENVIRONMENT CONCEPT

### 3.1 Overview

Figure 1 shows the developed concept of a test environment visualized in 3D. The test environment consists of five areas, where four of them are shown in Figure 1:

- Laboratory test bench
- Test hall
- Gate-based test setup
- Cart-based test setup
- Open field (not visualized)

From top to bottom, the similarity to real-world conditions increases. In addition, the level of control over environmental conditions, such as lighting conditions or the amount and distribution of dust, decreases from top to bottom. The test environment consists of an indoor and an outdoor area. Tests can be carried out indoors all year round. This means that the tests are not dependent on weather conditions, as outdoors. In the laboratory setup, either a camera or a stereo camera can be used. In addition to cameras and stereo cameras, the other test setups also allow the evaluation of LiDAR and RADAR sensors.
In the laboratory test bench, tests can be carried out in a defined environment on a laboratory scale. A camera or stereo camera is mounted on one side of a chamber, and a monitor is on the other. The second test setup indoors is a large-scale test hall. Realistic standardized test objects can be placed in the hall. In particular, a dummy of a person standardized (ISO 19206-2, 2018) and a test object standardized (ISO 18497-4, 2024) should be used for reproducible testing. ISO 19206-2 (2018) defines the characteristics of a person dummy, such as its size, so that the sensor output matches that of a real human being. The standardized dummy of a person is called a dummy in this work. ISO 18497-4 (2024) is a barrel with two diameters, the larger diameter at the bottom and the smaller diameter at the top. The test object is painted matt olive and is intended to represent a small person or animal. The lighting conditions and the amount and density of dust can be controlled in both setups. This means that all variables can be controlled. Both test setups enable recording reference-based sensor data with and without dust.

During tillage, it is challenging to record reference-based data. This issue is addressed by the outdoor test setups.

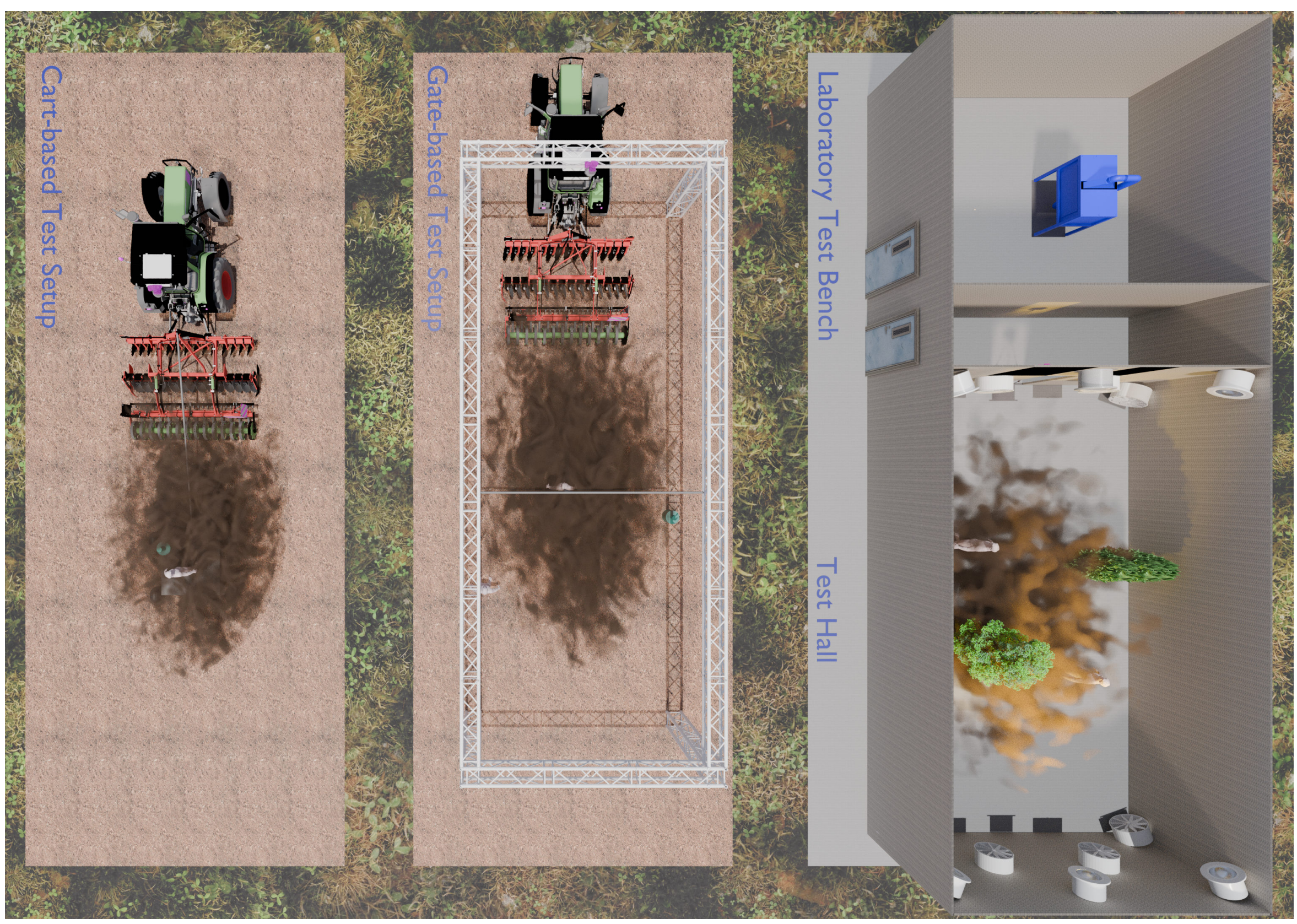

Fig. 1. Top-down view of the designed indoor and outdoor environment. The test environment consists of four different areas. The indoor area features a laboratory test bench designed explicitly for camera validation. The indoor area also includes a test chamber, a larger-scale space featuring fans and lights to raise and light up dust. The outdoor area is divided into three areas for field tests. The gate-based area consists of an aluminum traverse with four pillars through which a tractor can drive, and sensors mounted to it. The cart-based test field is a tractor-pulled carrier where test objects are placed on a cart. Finally, the open test field is for tests during regular agricultural work. The open test field is not illustrated for simplicity reasons.

The gate (see Figure 1) consists of an aluminum traverse with a rail system, where sensors can be mounted. Standardized test objects can be attached to the rail system. The tractor drives through the gate during soil cultivation, allowing sensor data to be recorded with and without dust. This enables thorough validation of the sensors and algorithms. Furthermore, a cart-based setup is designed, where the cart is attached to the implement via a cable. The distance can be adjusted using the length of the cable. Test objects can be placed on the cart. Thus, test objects are located in the center of the dust cloud. More importantly, the cart-based setup allows data to be recorded to a greater extent. Finally, an open field is essential for a test environment, as it allows real-life conditions to be investigated. The field must therefore be available for agricultural work all year round for testing purposes.

The individual test setups are explained in detail below.

### *3.2 Indoor Test Environment*

*Laboratory Test Bench* The laboratory test setup is shown in Figure 2. It consists of a chamber with glass walls at the front and sides for monitoring. A cone is attached to the top and bottom, which are connected via an external ventilation pipe. The cone, which runs across the entire floor, prevents dust from settling on the floor, for example, in the corners. A fan is integrated into the ventilation pipe. The fan regulates the air flow and thus the dust distribution. Dust can be added or removed via a maintenance flap. The camera is mounted on a sliding frame on one side of the box. The sliding frame adjusts the camera position to suit the lens and camera. A display is integrated into the wall opposite to the camera. Various images can be shown on the display. This allows multiple backgrounds to be displayed, such as fields at the forest's edge, near residential areas, or even in different countries. The display allows viewing open-source images from the web or dust-free images previously captured in the field. In addition, LED strips for various lighting scenarios are integrated into the setup. The lighting must match the brightness of the sun visually. Overall, the most important parameters, such as the lighting conditions, dust density, and the dust distribution, are controllable. During data acquisition, a camera with a lens is first focused on the monitor. An image is then displayed on the screen. A picture is captured from the displayed image. The fan is then started, which distributes the dust in the chamber. Another image is recorded with the dust in the chamber. Next, the fan is switched off, a new image is displayed, and the process is repeated. The test setups allow running a program with different lighting conditions and fan speeds. This makes it possible to capture an extensive dataset with a high degree of diversity in a laboratory environment. A stereo camera can also be used. For data collection small-scale objects can be placed inside the chamber on a steel grid.

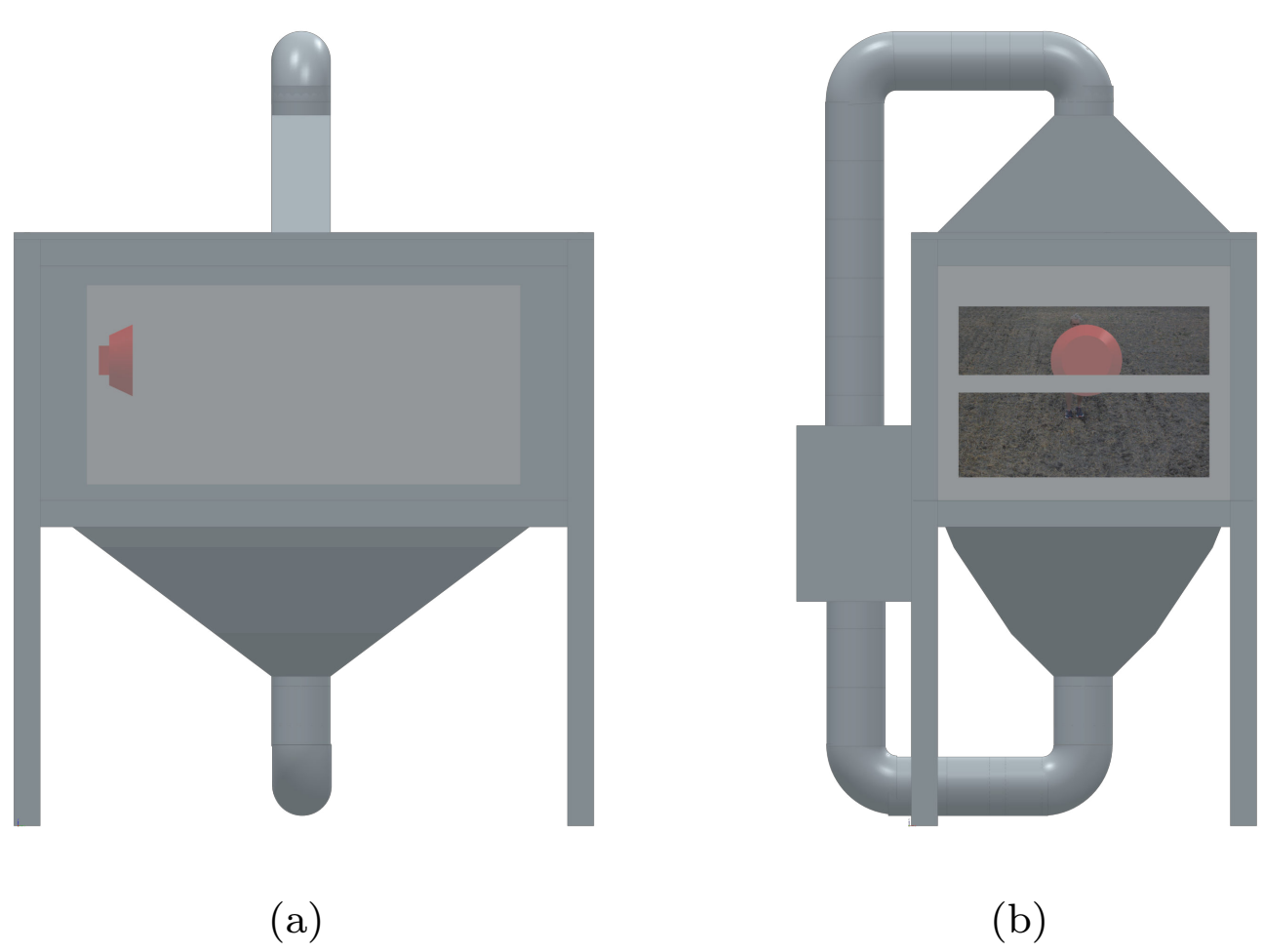

(a) (b)

Fig. 2. Design drawing of the laboratory test bench. The system consists of a chamber with a camera on one side and a monitor on the opposite side, which can display various backgrounds. The camera is illustrated with a red camera symbol, and an image is displayed on the screen. The dust is circulated using a ventilation system. As a result, the dust density can be controlled with a fan. (a) Front view. (b) Side view.

*Test Hall* The lab test rig can only be used for camera or stereo camera systems due to its setup. This is where the test hall comes in. The test hall is a large room in which data can be recorded. The hall is equipped with fans to stir up dust and lamps that visually resemble the sun's illumination. The sensors are located in a wall cutout so that they are not directly exposed to dust and are easily accessible. Fans blow the dust away from the sensors. The size of the hall makes it possible to place real-life test objects such as bushes, hedges, and standardized test items.

The test objects are placed at different distances from the sensor system. The amount and distribution of dust can be controlled via the fans. The lighting can be used to simulate various lighting conditions during the day. First, images of the dust-free scene are recorded. Next, a program is launched that controls the fan speed and lighting. The automatic program allows data to be recorded with different dust distributions under changing lighting conditions. Figure 3 illustrates different Lighting scenarios with dust raised in the test hall.

### *3.3 Outdoor Test Environment*

*Gate-based Test Setup* Figure 4 shows the proposed gate for reference-based data recording of dust raised during tillage or seeding. The gate with four pillars consists of aluminum traverses that are both torsion-resistant and lightweight. Additionally, the gate features a rail system where objects can be attached. The rail system allows dummies to be mounted to hover just above the soil.

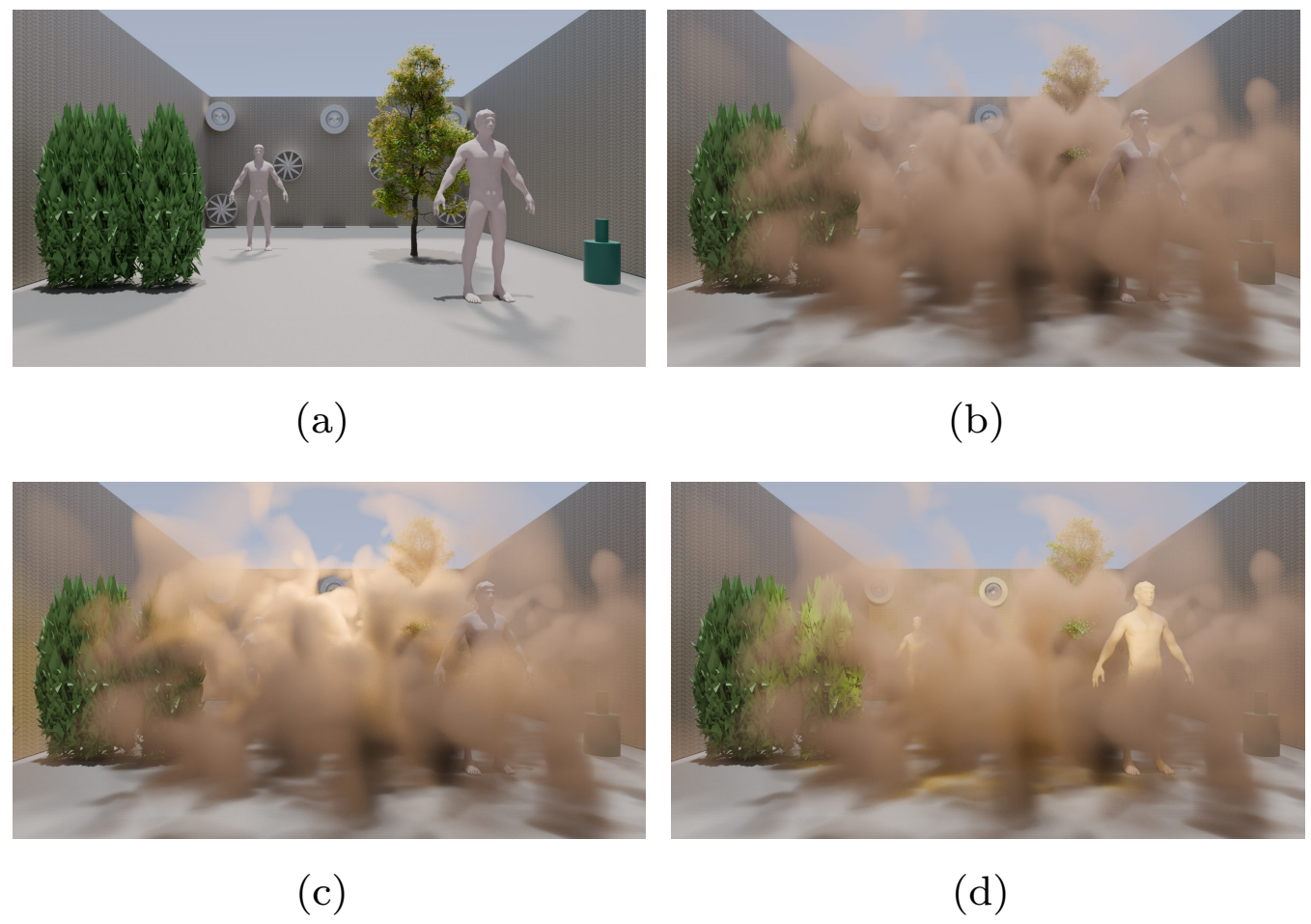

(a) (b)

(c) (d)

Fig. 3. Various views of the test chamber. The test chamber is a large-scale room equipped with sun-like lighting and fans in various locations. The chamber has interior fittings like trees, hedges, and certified dummies. Image from the sensor's perspective. (a) Dust-free image with all lights off. (b) Dusty scene with no lights. (c) Dust raised and lights opposite the sensors are switched on. (d) Dusty and the lights above the sensor are lit.

Furthermore, standardized test objects can be placed either to the left or right of the tillage track. Therefore, the gate must be high and wide enough to allow agricultural machinery to pass through safely. Sensors are installed on both sides of the gate and at the top of it. The sensors must be aligned to point at an angle downwards to record the dust raised during tillage. It is essential to choose the right lens for the camera. Ideally, the field of view should correspond to the percentage of dust visible in images recorded from the tractor cab. In addition, vibrations should be monitored during testing to investigate the influence on data acquisition in more detail.

The test procedure is the following:

(1) The tractor cultivates the field and thus raises dust.
(2) The person mounted to the rail system is waiting at one corner of the gate.
(3) The tractor enters the Gate while cultivating the field.
(4) The rail-mounted dummy moves into the middle after the tractor passes the entry.
(5) While the tractor passes through the traverse, the dummy follows the tractor via the rail system.
(6) The tractor leaves the gate while the dummy remains at the last position.
(7) The dummy returns to its starting position.

This allows data to be recorded while the dummy moves and changes its position and distance from the sensors.

*Cart-based Test Setup* During soil cultivation, a person can follow directly behind the implement in the middle of the dust cloud, which is a safety risk. The dust can be so dense that the person is not visible, which poses a significant problem for subsequent labeling. One solution is recording data where a dummy is standing next to the track of the working tractor. However, due to external influences such as wind, the person may only be partially obscured in dust. As a result, tests may have to be repeated several times to record the best possible data.

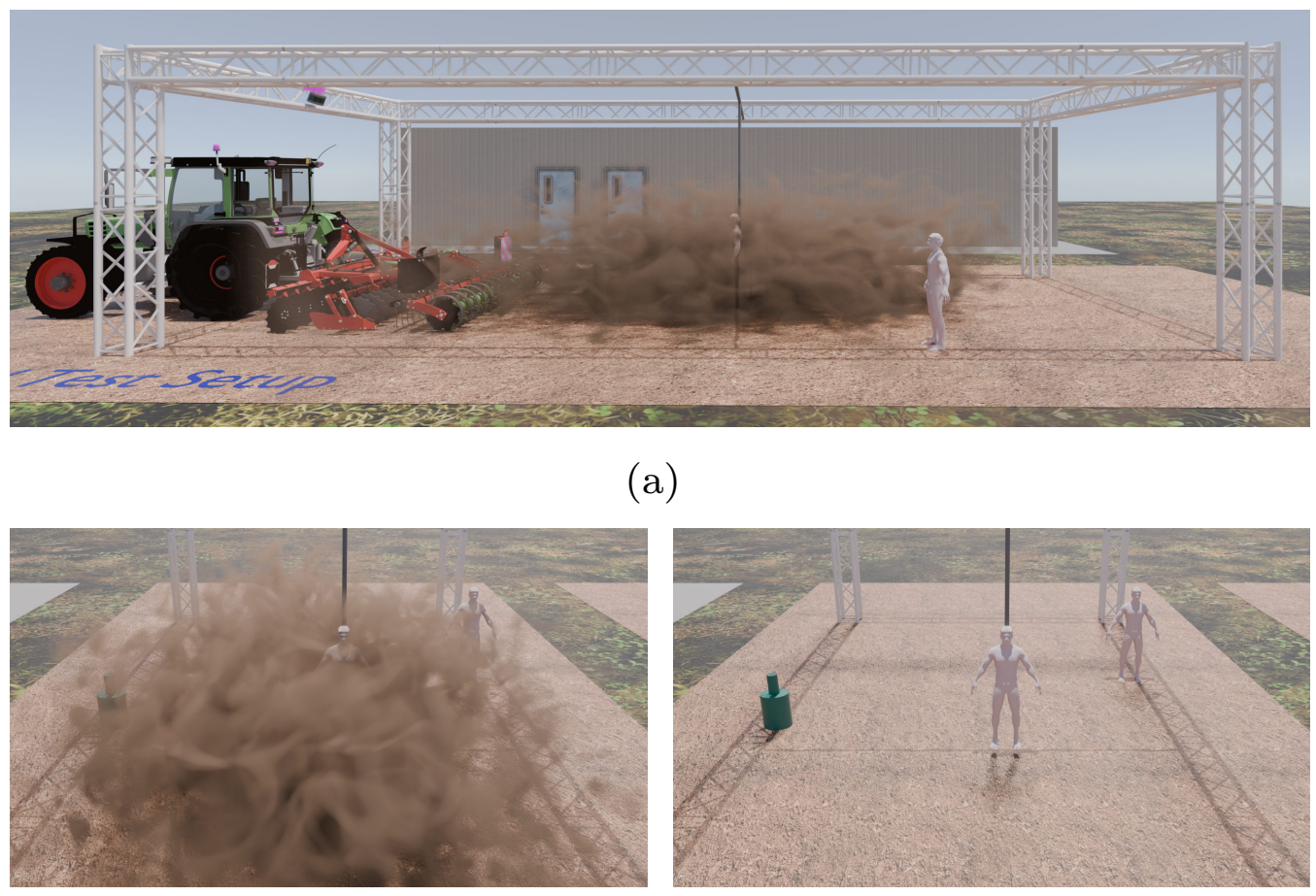

(a)

(b) (c)

Fig. 4. Overview of the outdoor test field with gate for mounting of sensors. A tractor can drive through the gate to cultivate the field. This allows dust-free and dusty sensor data to be recorded. (a) Side view. (b) Sensor view with dust; (c) Sensor view without dust.

The cart-based approach solves the aforementioned problems. Figure 5 illustrates the concept. The concept is based on a cart attached to the tractor's implement via a cable. Standardized test objects can be placed onto the cart. The cart is designed to be flat so that the test objects are as close to the ground as possible. The length and thus the distance to the tractor can be adjusted using the cable. This allows the dummy to be located precisely, even in dense dust. Thus, the setup enables a detailed validation of person detection and provides conclusions about how performance varies with the distance between the dummy and the sensor system. The domain gap is minimal since the tests are conducted during soil cultivation. It can therefore be assumed that the results can be transferred almost directly to real-world applications.

*Open Test Field* The test environment also includes a test field that can be used for unrestricted testing. The test field should be ideally available all year round and uncultivated. This makes it possible to carry out any field work as soon as the soil is dry enough. Flat-working implements, such as a disc harrow or a seed drill, are suitable for stirring up dust. During the tests, people can move freely around the field. This allows testing the environmental perception of autonomous agricultural machinery.

## 4. RESULTS AND DISCUSSION

Table 1 compares the previously introduced test setups with some of the most relevant characteristics. Both indoor benches and the gate-based setup can be used to record reference-based data. By using the cart-based and the open test field, it is only possible to record reference-less data, without matching dusty and dust-free images. Excluding the laboratory test bench, standardized test objects can be placed anywhere in the test environment. However, dust-free images with a test object can be displayed on the screen in the laboratory test bench. Recording a large-scale dataset with high variance is easily achieved on the open test field. The cost of data acquisition in the test hall and the gate-based system is significantly higher because the test objects are manually rearranged. The dust density can be controlled by adjusting the fan speed of the lab test bench and the test hall. Furthermore, the impact of wind direction in the test hall can be examined by regulating the different fans. Both indoor test setups allow the control of lighting conditions using artificial lighting.
Each test setup is intended to investigate the effects of dust on sensor perception. However, they differ in how they can be used. The laboratory bench can be used to develop stereo camera and single camera-based algorithms to remove the dust, since the setup allows recording enough data to, e.g., train a neural network. While the other test setups additionally enable the analysis of LiDAR, RADAR, etc. The test hall is intended for year-round validation and testing. Additionally, test objects can be set up at defined distances from the sensors. On the other hand, the outdoor gate-based setup enables the validation of dust removal and person detection methods where test objects stand in real-world dust clouds during tillage. The cart-based setup is intended to develop and validate person detection algorithms with a person in the middle of the dust cloud, representing the worst-case scenario. All previously developed and validated hardware and software must be tested in the field during real-world use. This happens on the open test field.

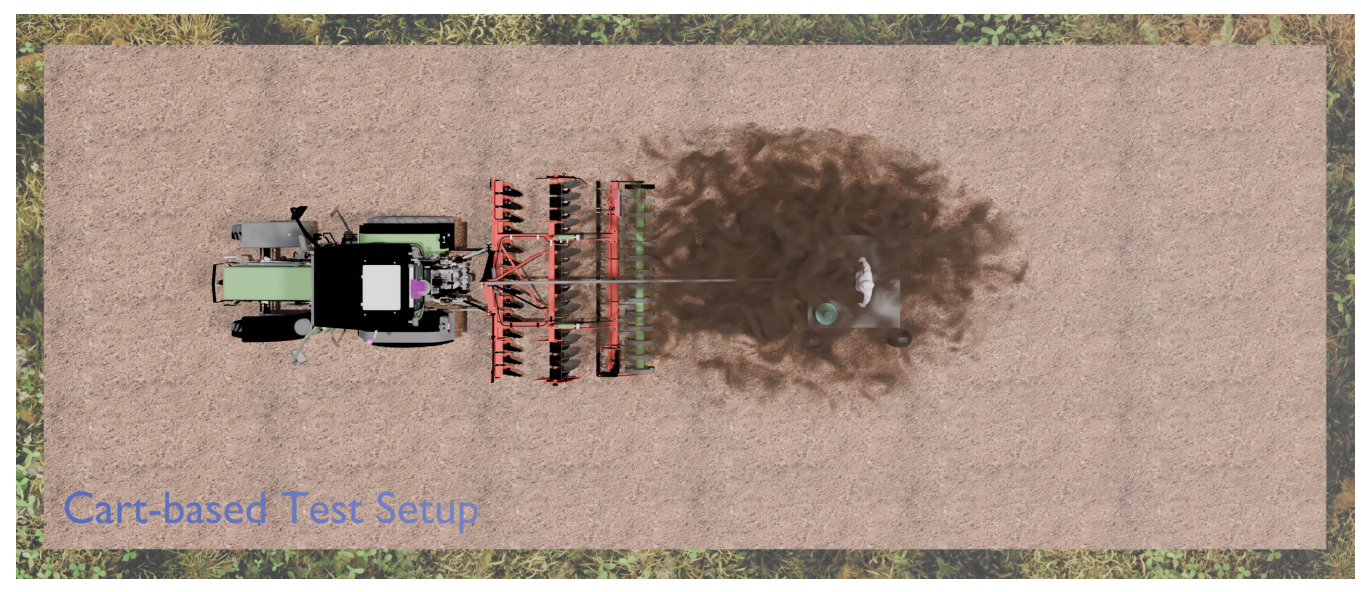

(a)

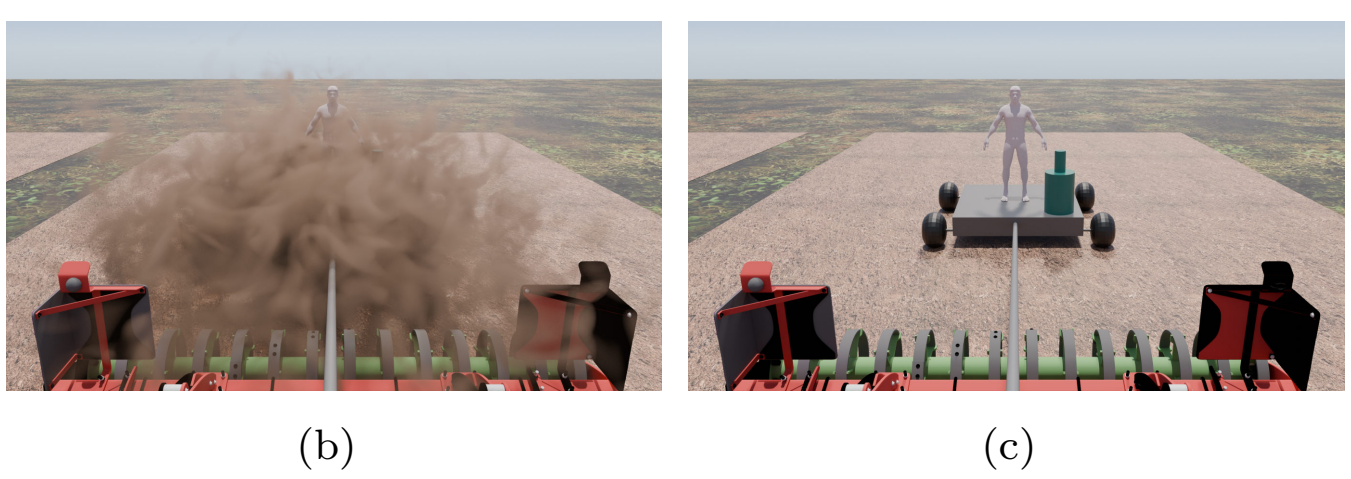

(b) (c)

Fig. 5. Overview of the cart-based test setup with a cart attached to a tractor implement using a wire rope. Certified test objects can be mounted on the carrier. (a) Top-down view. View from the perspective of the sensors mounted to the tractor's rear window with dust (b) and without dust (c)

Implementing the described test setups requires a significant amount of work and investment. The laboratory test bench and the cart-based test setup require less effort to setup and operate than the test hall and the gate-based setup. In summary, each of the test benches developed serves a different purpose.

Table 1. Comparison of the proposed test setups and the defined properties. ✓is used if a condition is fulfilled and ✗if not. ↑ and ↓ indicate if the cost or level of control is high or low.

| | | Laboratory test bench | Test hall | Gate-based | Cart-based | Open field |
|---|---|---|---|---|---|---|
| Recording of reference-based data | | ✓ | ✓ | ✓ | ✗ | ✗ |
| Recording of reference-less data | | ✗ | ✗ | ✗ | ✓ | ✓ |
| Use of standardized test objects | | ✗ | ✓ | ✓ | ✓ | ✓ |
| Time and effort required to collect data | | ↓ | ↑ | ↑ | ↓ | ↓ |
| Sensors | Stereo camera & camera | ✓ | ✓ | ✓ | ✓ | ✓ |
| | LiDAR, RADAR, etc. | ✗ | ✓ | ✓ | ✓ | ✓ |
| Level of control over the dust density | | ↑ | ↑ | ↓ | ↓ | ↓ |
| Dust distribution similar to real-world | | ✗ | ✗ | ✓ | ✓ | ✓ |
| Level of control over lighting conditions | | ↑ | ↑ | ↓ | ↓ | ↓ |
| Various lighting conditions | Day | ✓ | ✓ | ✓ | ✓ | ✓ |
| | Night | ✗ | ✓ | ✓ | ✓ | ✓ |

## 5. SUMMARY AND CONCLUSION

Existing environmental test benches are either entirely or partially unsuitable for investigating the effects of dust on environmental perception and sensors. Dust-free reference data must be reproduced without deviation to investigate dust's effects on sensor perception. This results in special requirements for a test bench. The extent to which the results can be transferred to real-world applications is a significant challenge for all test benches.

For this purpose, indoor and outdoor test benches are proposed. The indoor test setups enable year-round testing to develop sensors and algorithms. The outdoor setups allow the validation close to real-life conditions during soil cultivation. The setup meets the predefined requirements, allowing reproducible data recording. Furthermore, standardized testing objects can be used for validation. Through the combination of indoor and outdoor test setups, sensors and algorithms can be tested in scenarios that reflect the real-world dust distribution and lighting conditions. In addition, our results show that each test setup serves a different purpose.

Overall, a test facility with test benches modeled in three dimensions is proposed in this work for sensor and algorithm validation in dusty agricultural conditions.

## ACKNOWLEDGEMENTS

The Zeppelin Foundation funded Peter Buckel